# A Novel Approach for Deterioration and Damage Identification in Building Structures Based on Stockwell-Transform and Deep Convolutional Neural Network

Vahid Reza Gharehbaghi[a], Hashem Kalbkhani[b], Ehsan Noroozinejad Farsangi[c]*, T.Y. Yang[d], Andy Nguyen[e], Seyedali Mirjalili[f,g] & C. Málaga-Chuquitaype[h]

[a] Research Scholar, Kharazmi University, Faculty of Civil Engineering, Tehran, Iran (vahidrqa@gmail.com)
[b] A/Professor, Department of Electrical Engineering, Urmia University of Technology, Urmia, Iran (h.kalbkhaniuut.ac.ir)
[c] A/Professor, Faculty of Civil and Surveying Engineering, Graduate University of Advanced Technology, Kerman, Iran (***Corresponding Author**: noroozinejad@kgut.ac.ir)
[d] Professor, The University of British Columbia, Vancouver, Canada (yang@civil.ubc.ca)
[e] A/Professor, University of Southern Queensland, Springfield Campus, Queensland 4030, Australia (Andy.Nguyen@usq.edu.au)
[f] Center for Artificial Intelligence Research and Optimization, Torrens University Australia, Fortitude Valley, Brisbane, QLD 4006, Australia (ali.mirjalili@laureate.edu.au)
[g] Yonsei Frontier Lab, Yonsei University, Seoul, Korea
[h] A/Professor, Department of Civil and Environmental Engineering, Imperial College London, UK (c.malaga@imperial.ac.uk)

## Abstract

In this paper, a novel deterioration and damage identification procedure (DIP) is presented and applied to building models. The challenge associated with applications on these types of structures is related to the strong correlation of responses, which gets further complicated when coping with real ambient vibrations with high levels of noise. Thus, a DIP is designed utilizing low-cost ambient vibrations to analyze the acceleration responses using the Stockwell transform (ST) to generate spectrograms. Subsequently, the ST outputs become the input of two series of Convolutional Neural Networks (CNNs) established for identifying deterioration and damage to the building models. To the best of our knowledge, this is the first time that both damage and deterioration are evaluated on building models through a combination of ST and CNN with high accuracy.

**Keywords:** Deterioration, Damage, Stockwell Transform, Convolutional Neural Networks (CNNs), Deep Learning

**Introduction and Motivations**

The accumulation of deterioration in structures during their lifetime leads to the reduction of strength and performance and imperils their serviceability and safety (Monavari, 2019). Subsequently, the capability of a structure to endure extreme incidents diminishes over time. Hence, preserving the functionality of structures and enhancing their performance level to reduce upkeep costs have become a central focus in structural health monitoring (SHM). It is noteworthy that embedding an SHM system in the structure makes it possible to continuously monitor the system's changes during operation time. This potential allows determining the appropriate moment for essential or preventive maintenance action (EMA and PMA) and thereby reduces the risk level of probable breakdowns (Figure 1).

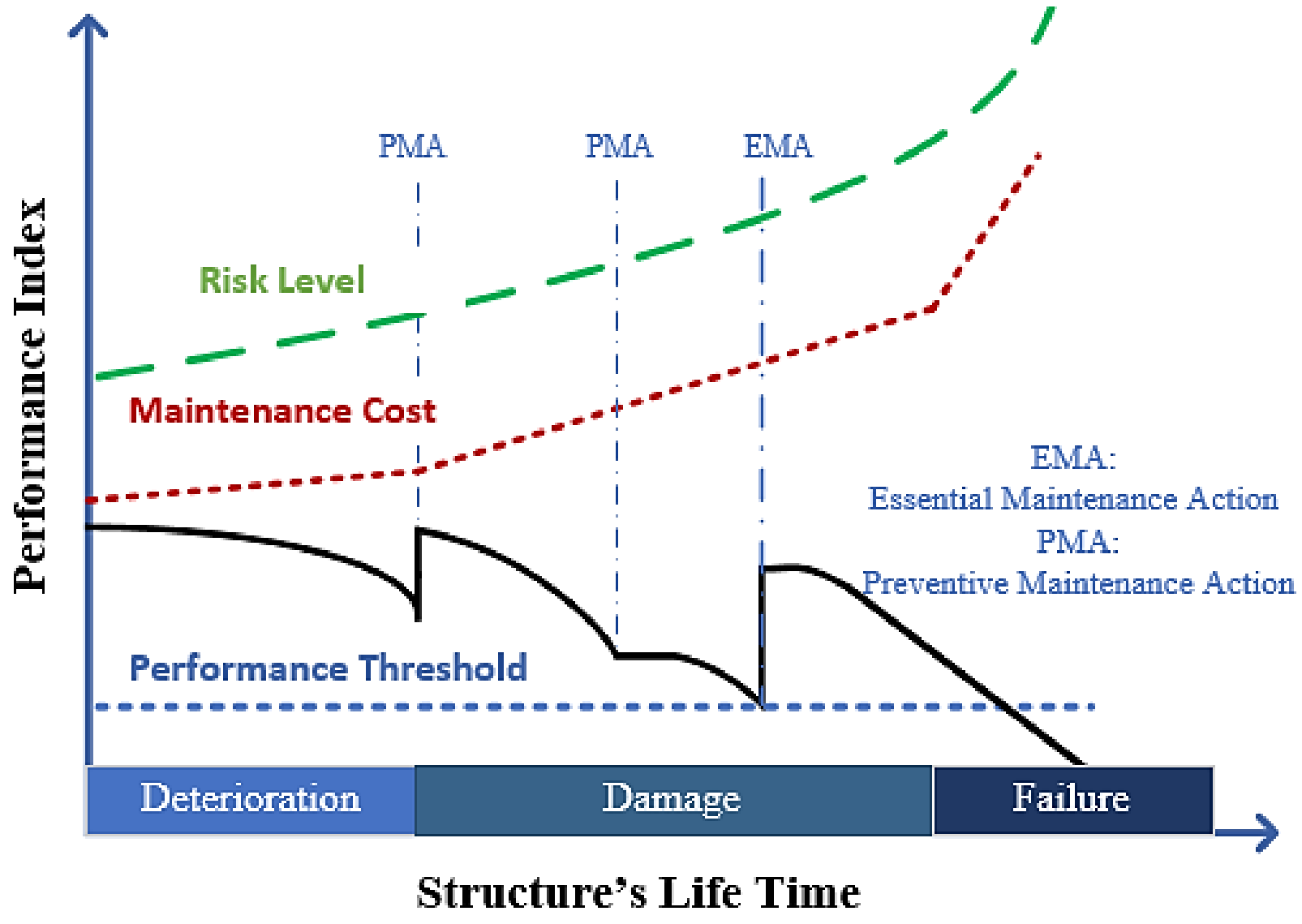

*Figure 1. Effect of maintenance actions on structure [adapted from (Berkowski & Kosior-Kazberuk, 2017)]*

Based on a survey conducted on 225 building cases in the USA, it has been demonstrated that deterioration is a frequent concern deserving attention (Wardhana & Hadipriono, 2003). Therefore, identifying structures' responses to detect variations is supposed to be a significant task that can ensure safety and related financial considerations.

Damage and deterioration are defined as degradation in a system's performance due to changes in material, components, or structure connections. However, compared to damage, deterioration identification requires more accurate and reliable techniques (Monavari, 2019). Recently, several studies have focused on damage identification through variations in the structure responses.

Nonetheless, while these surveys have illuminated numerous practical approaches in the sense of identification and localization of damage on simple components or bridges (Gillich et al., 2017; Sha et al., 2020) (Jayasundara et al., 2020), fewer have been done to assess slight damages on building structures (Regier & Hoult, 2015). This concern is exacerbated in complex structures like buildings, where a strong correlation of responses is considered one of the challenging issues (Gharehbaghi et al., 2020). Deterioration recognition in complex systems requires laborious efforts since not only ambient excitations lead to lower amplitudes in responses, but also noise and operational effects influence the performance of detection.

For the sake of illustration, in Figure 2, two samples of signals corresponding to a damaged and deteriorated state are plotted versus the baseline or healthy state. The signals are captured from two building specimens with distinct deterioration and damage scenarios discussed in the following sections. As can be seen, the deterioration signal is hardly distinguished from the healthy signal. By contrast, the signal of damage reveals significant variations compared to the baseline state and thereby is readily apparent even to the naked eye.

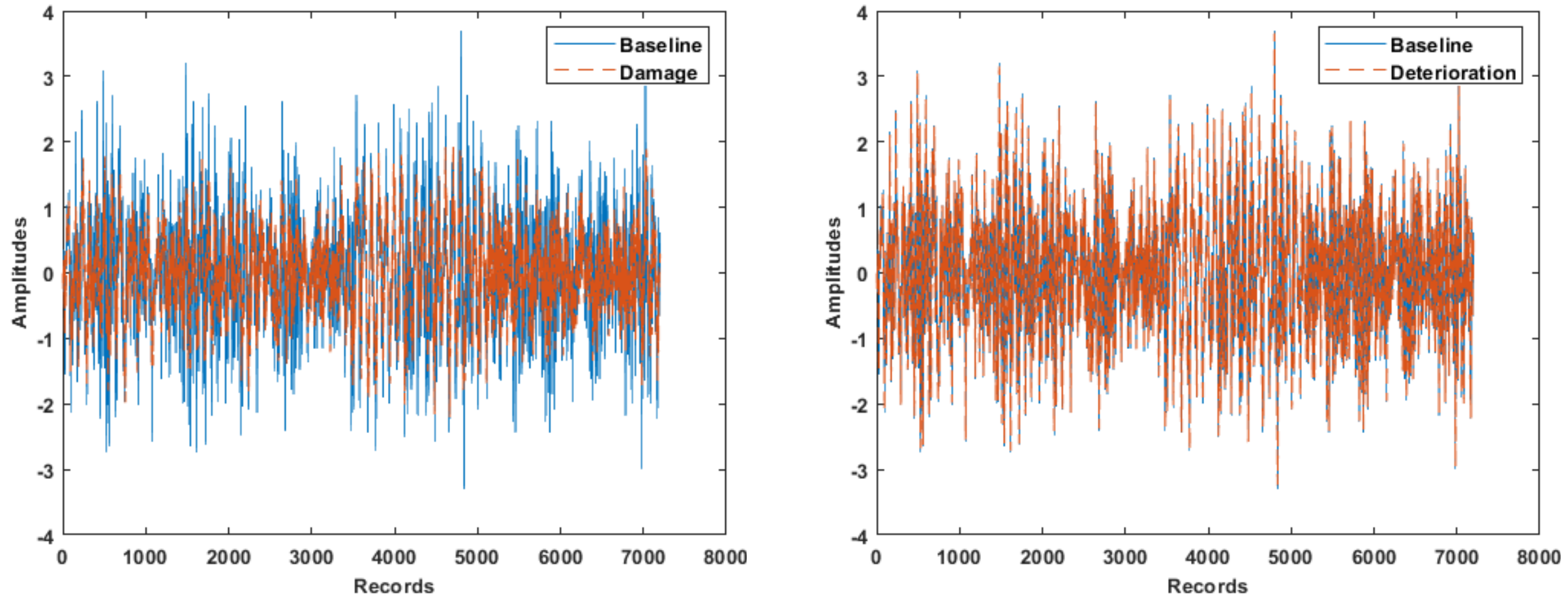

(a) Signal of damage vs. baseline (b) Signal of deterioration vs. baseline

*Figure 2. Comparison of deterioration and damage signals*

There are two main approaches to conduct SHM in practice, including model-based and data-driven SHM. In the first technique, a complicated model of the monitored system is needed, which requires information regarding geometry, materials, and boundary conditions of the structure. In the former, however, the real data collected from sensors embedded in the structure is deployed to recognize various system conditions. Stochastic methods coupled with signal processing techniques are the principal tools for disclosing hidden features within the responses in the realm of data-driven SHM. Herein, for the sake of pattern recognition, machine learning algorithms are prevalent, including classical machine learning algorithms such as decision trees, Support vector machines (SVMs), and K nearest neighborhoods (KNNs) or deep learning approaches.

According to the above limitations and motivations, this article attempts to establish a system for deterioration assessment in a building based on a data-driven strategy and deep learning for pattern recognition as one of the most promising trends in Artificial intelligence. The rest of the paper is organized as follows.

First, some of the data-driven studies with an emphasis on deterioration detection are presented. Next, deep learning basics are introduced, and relevant investigations are reviewed. Third, the deterioration identification procedure (DIP) is illustrated in sequential steps. Fourth, the specimens for the deterioration and damage study are depicted. Last, the DIP is applied to the case studies, and the results are discussed.

**A review on deterioration detection in building structures**

Signal processing, stochastic methods, and machine learning algorithms play a significant role in extracting features, developing damage indices, and pattern recognition in data-driven SHM. Time series, Fourier transformation, wavelets, Wigner-Ville distribution (WVD), empirical mode decomposition (EMD), and Stockwell transform are some of the signal processing used frequently by researchers. Herein, some studies that attempted to identify slight damages on building structures are reviewed.

Santos et al. (Santos et al., 2016) were successful in detecting damages on a real bridge using unsupervised machine learning methods. To this end, Neural Networks were adopted for structural estimation response, and clustering techniques were applied for classifying the neural network estimation errors. A moving windows process method was utilized for enduring the continuous online procedure. Consequently, small reductions in single cables' stiffness were detected using a small number of inexpensive sensors.

Wang et al. (Wang et al., 2016) deployed the Autoregressive (AR) time series coefficients for detecting slight damages on a cantilever beam and a building SHM benchmark. It was deducted that the control charts had the potential to identify damage in the early stages, even in a noisy environment. However, the precise location and severity of damage could not be indicated based on that study.

Monavari et al. (Monavari et al., 2018) developed a signal-based approach utilizing residuals of AR time-series. The potential of time series for analyzing ambient vibrations and their sensitivity and reliability compared to other vibration-based damage detection was the primary motive. In this study, a novel model order estimation algorithm was established to fit structural responses with more than 95% accuracy. Eventually, the statistical hypotheses of a two-sample f-test were applied to the AR model's residuals as a deterioration indicator. The efficiency of the proposed methodology was validated by detecting multiple deterioration locations on two multi-story RC frames in a noisy environment.

A robust model-based technique based on long-term monitoring data was proposed by Nguyen et al. (Nguyen et al., 2019). To build an empirical model of the complex structure, operational modal analysis (OMA) is used coupled with an initial FEM of the structure made by as-built drawings. In this paper, a hybrid model updating procedure was exploited, coupled with sensitivity analysis. The gradual, subtle area-section reduction was considered as the simulation of slight damage in life-cycle assessment. Subsequently, the authors efficiently predicted the deterioration mechanism under serviceability loading in a real existing case study of a 10-story benchmark structure.

Gharehbaghi et al. (Gharehbaghi et al., 2020) established a novel algorithm to select sensitive uncorrelated features derived from AR models and statistical indices. The selected features were then used for generating a stochastic pattern for each state in the system. A linear discriminant classifier was applied to the pattern for separating damage and deterioration conditions in two building models. It was shown that the proposed strategy had the potential for classifying different states of structure even in a noisy environment.

In another paper, Gharehbaghi et al. (Gharehbaghi et al., 2021) utilized time-frequency processing using wavelets coupled with some statistical insides for extracting features of damage and deterioration in specimens. A novel feature selection method extracted sensitive features based on baseline signals which built structural behavior patterns. Classical machine learning algorithms including ANN, SVM, and KNN were deployed for discriminating healthy and unknown status within building models. The efficiency of the proposed technique was verified under various conditions such as signal length, sampling rate, and the number of features. It was proven that the method could detect, locate, and assess the severity of damage and deterioration under environmental variabilities such as high noise and operational effects.

**Deep learning fundamentals**

In real-world intelligence applications, many factors and variations exist in the data. Extracting high-level and abstract features, such as detecting specific objects in a picture or finding individual pixels in a color image, is challenging. Many of these factors of variation require a cutting-edge, nearly human-level knowledge of the data. Deep learning addresses this issue in representation learning by decomposing a complex problem into more straightforward representations. This matter allows a computer to develop sophisticated models from simple concepts (Heaton, 2018). For instance, a deep leading algorithm can understand complex features for detecting unique objects within a picture using corners and contours of pixels.

A central feature of AI is the use of various Artificial Neural Networks along with some connected layers. It shares similar algorithms with Machine Learning, including four stages in terms of building a paradigm. Deep Learning is different from Neural Networks in some aspects, as demonstrated in Figure 3. As seen, the feature extraction and classification are conducted within a more complicated neural network. Notably, deep neural networks reveal higher accuracy and efficiency in big data than classical machine learning algorithms due to the high complexity of structure (LeCun et al., 2015).

Concerning configuration, deep neural networks are known via four major architectures, namely Unsupervised Pre-trained Networks (UPNs), Convolutional Neural Networks (CNNs), Recurrent Neural Networks, and Recursive Neural Networks (RNNs) (Patterson & Gibson, 2017). In brief, a UPN manages and also learns the input data in the network by itself. It weighs the first layer through unsupervised pre-trained data. Recurrent Neural Networks and RNNs are exploited to process time-sequential data. In this category, Long short-term memory (LSTM) (Luo et al., 2018) is one of the popular recurrent neural networks used for predicting the time sequences over time (Deng et al., 2020).

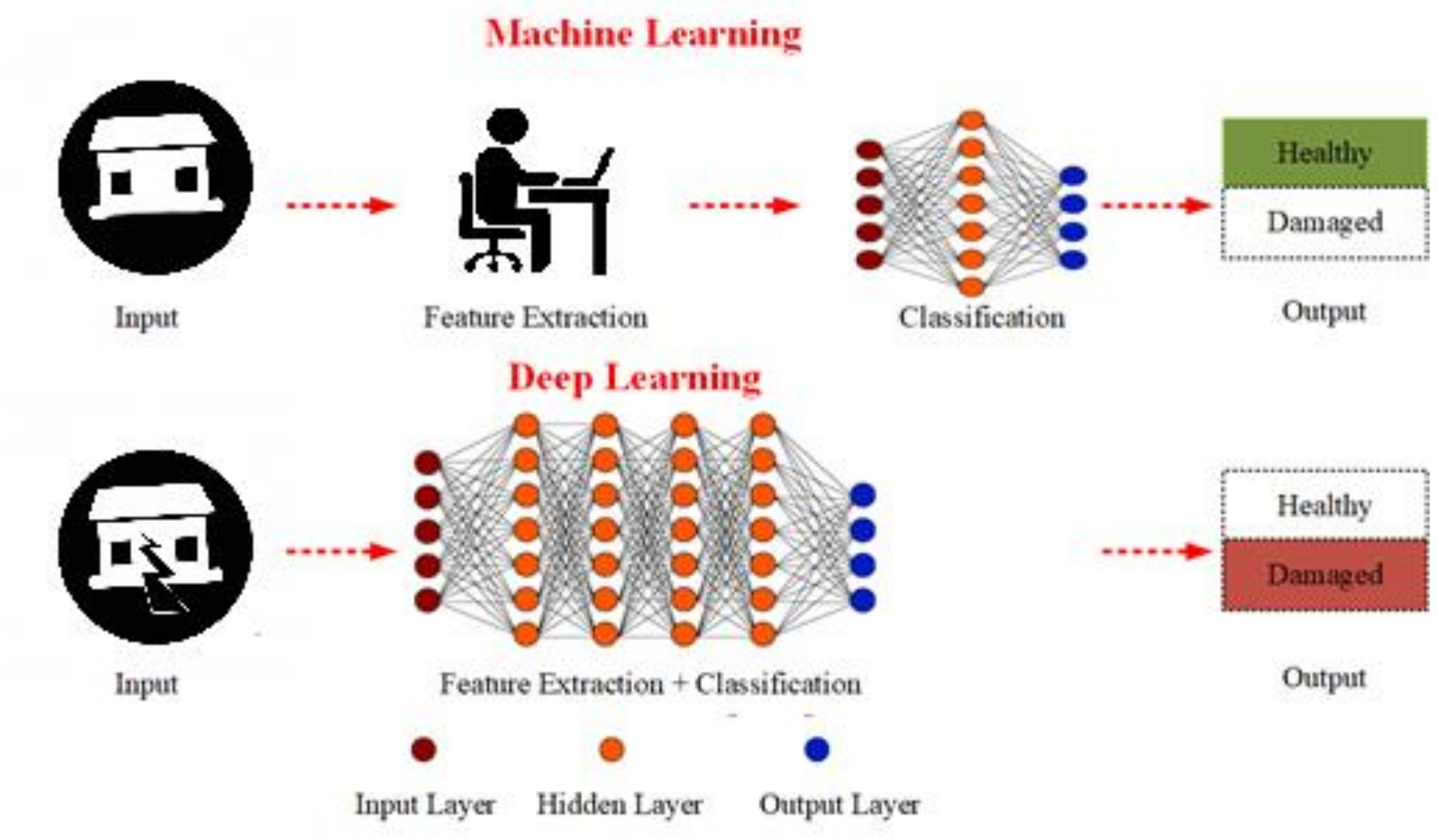

*Figure 3. Machine Learning vs. Deep Learning*

Compared to other types of networks, CNNs have been widely used in data science, especially for addressing image classification problems since they can learn directly from the raw images. They are used to extract the multi-scale localized spatial features from the input image by performing local filtering of the input image (Toh & Park, 2020).

CNNs consist of different layers, including convolution, rectified linear unit (ReLU), pooling, fully connected layers, and softmax, as shown in Figure 4. The first layer obtains high-level features from a part of the data using small kernels. For the sake of fast training and having lower sensitivity to initialization, the convolution's output is usually normalized using mini-batches.

The ReLU is a nonlinear activation function generating new outputs as follows:

$$x_{out} = \max\{0, x_{in}\} \tag{1}$$

where $x_{in}$ depicts the output elements of the batch normalization layer and $x_{out}$ shows the output of the ReLU layer. Downsampling is implemented on the output feature map of the ReLU layer.

Herein, max pooling is the most common pooling operation that uses the maximum value of a neighborhood area in the feature map. Additionally, in order to avoid overfitting errors, the dropout layer is used. In this layer, each neuron may be eliminated in a particular repetition of the training stage with the probability of $p$ . Notably, this layer is just employed in the training stage and is not used in validation or testing. Subsequently, the fully connected and softmax layer can classify the input data for identification purposes.

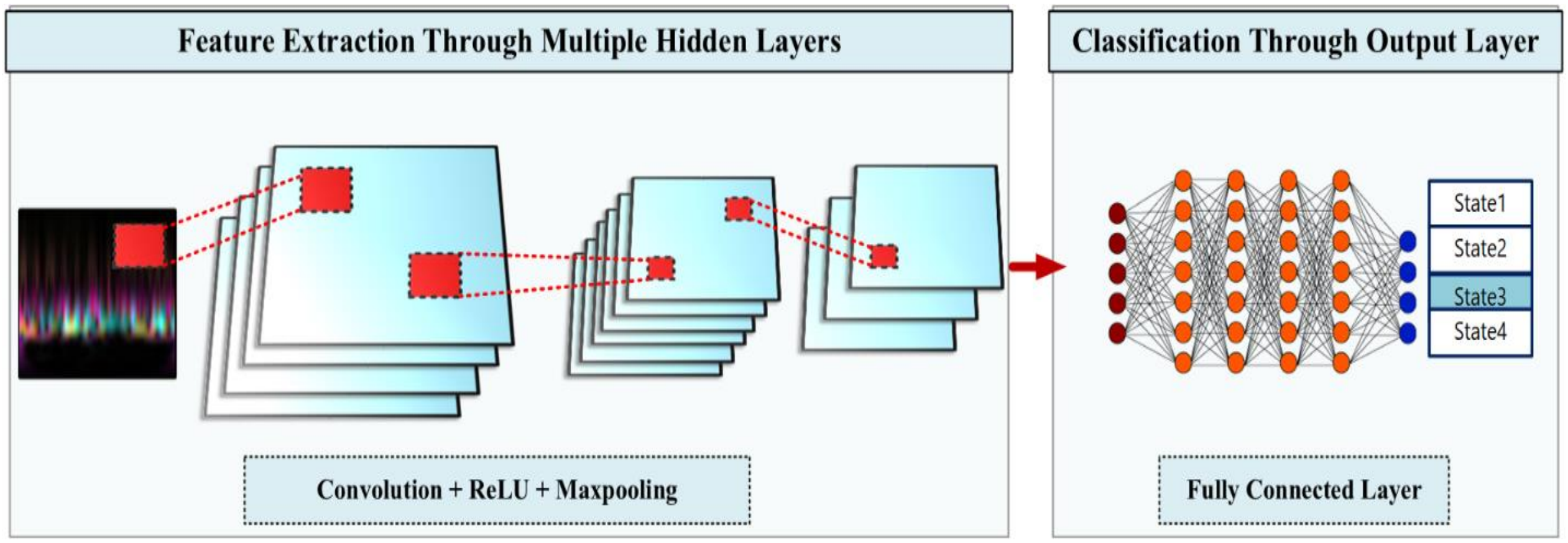

*Figure 4. Typical architecture of CNN (Patterson & Gibson, 2017)*

In general, a combination of SHM and deep learning can be investigated through two central frameworks. Image-based Deep SHM (ID-SHM) employs images as input, and Signal-based Deep SHM (SD-SHM) utilizing signals or one-dimensional data. Following the first framework, as the most common type, deep learning is employed instead of human sight to classify and analyze collections of raw images.

It is essential to note that sufficient numbers of images should be collected via various kinds of hardware as well as pieces of equipment, including digital cameras, infrared cameras, robotic systems, and Digital Image Correlation (DIC) systems. In light of this framework, multiple image processing operations are conducted to obtain information from images. This information is then employed as the input features of a deep learning network.

Regarding ID-SHM, Zhang et al. (Zhang et al., 2016) proposed a deep learn-based method for crack detection in pavements using a data set of 500 images captured with a smartphone camera. They trained the network through the Stochastic Gradient Descent (SGD) method accelerated by GPU rather than ReLU as the activation function. SVM and the Boosting algorithms were compared with the proposed method. As a result, it was shown that CNN had higher performance in classifying crack patches from the background.

Kim et al. (Kim et al., 2019) established a novel method for locating cracks on concrete surfaces in noisy patterns. In that study, image binarization was put to use to extract cracked regions coupled with a CNN for classification. The classification models were built based on speeded-up robust features (SURF) and CNN. It was concluded that the overall performance of the CNN method was better than the SURF-based method in the majority of cases. Moreover, combining deep neural networks with SVM classifiers with local/global features could enhance the accuracy compared to separate deployment of each method.

Li et al. (Li et al., 2020) studied automatic crack inspection within a tunnel. Accordingly, U-Net and a CNN were combined and an updated version of the clique (CliqueNet) was used to distinguish cracks from backgrounds using digital camera images. A total of 200 manually labeled images at a pixel level were used for the training and testing. The proposed approach was simple but effective, showing success in detecting internal cracks with high accuracy.

On the other hand, SD-SHM methods use structure responses (accelerations, displacements, velocities, and strains) to detect an anomaly. Based on that, signal processing techniques, namely time-series, Fourier transform (FT), Short-time Fourier transform (STFT), Wavelet Transform (WT), Stockwell transform (ST), and Empirical mode decomposition (EMD) are utilized to analyze the responses. By applying these operations to the recorded signal, numerous attributes within specific domains (time, frequency, or time-frequency) are extracted as sensitive features that can reflect changes in the structure.

In earlier approaches, conventional classifiers like SVM, Principal Component Analysis (PCA), and Artificial Neural Network (ANN) were employed to pre-process and classify data. Nonetheless, in SD-SHM, researchers benefit from promising deep learning networks as a powerful tool along with signal processing techniques. For example, Abdeljaber et al. (Abdeljaber et al., 2017) established an adaptive CNN that could detect different damage scenarios on a large-scale building model using acceleration measurements. Damages were simulated by loosening the bolts at the joint. Moreover, 31 damage scenarios were utilized for training the network. It was shown that the method could locate single and double damage cases in real-time.

Avci et al. (Avci et al., 2018) exploited the methodology proposed in (Abdeljaber et al., 2017) with a shallow structure, but this time with the aid of triaxial accelerations measured with a wireless sensor network (WSN). The authors attempted to find the most sensitive axis for detecting damage. Their results demonstrated that the modified approach was promising in terms of the location of damage under ambient vibrations.

Sarawgi et al. (Sarawgi et al., 2019) introduced an experimental procedure that was capable of detecting damage on an I-AISC benchmark structure through 4 phases with the aid of a CNN. It was shown that the proposed approach was capable of detecting and localizing damage in the laboratory model. Further studies regarding the application of CNN in SHM are reviewed in this reference (Avci et al., 2020).

The above review indicates that deep learning has become a popular tool for damage identification. However, its potential has not been explored and harnessed enough in the case of deterioration. In light of this, the authors present an intelligent system for deterioration identification in building structures.

## Case Studies

In this section, the case studies are presented in details.

### *First Case-Study: Deterioration Model*

In this paper, a full-scale finite element model (FEM) of a three-story reinforced concrete (RC) frame, which has been modeled in the IDARC program (Reinhorn et al., 2006), is used to confirm the efficiency of the proposed method. The height of each story and the bay width is 3000(mm) and 4000(mm), respectively. The element sections of the model are shown in Figure 5.

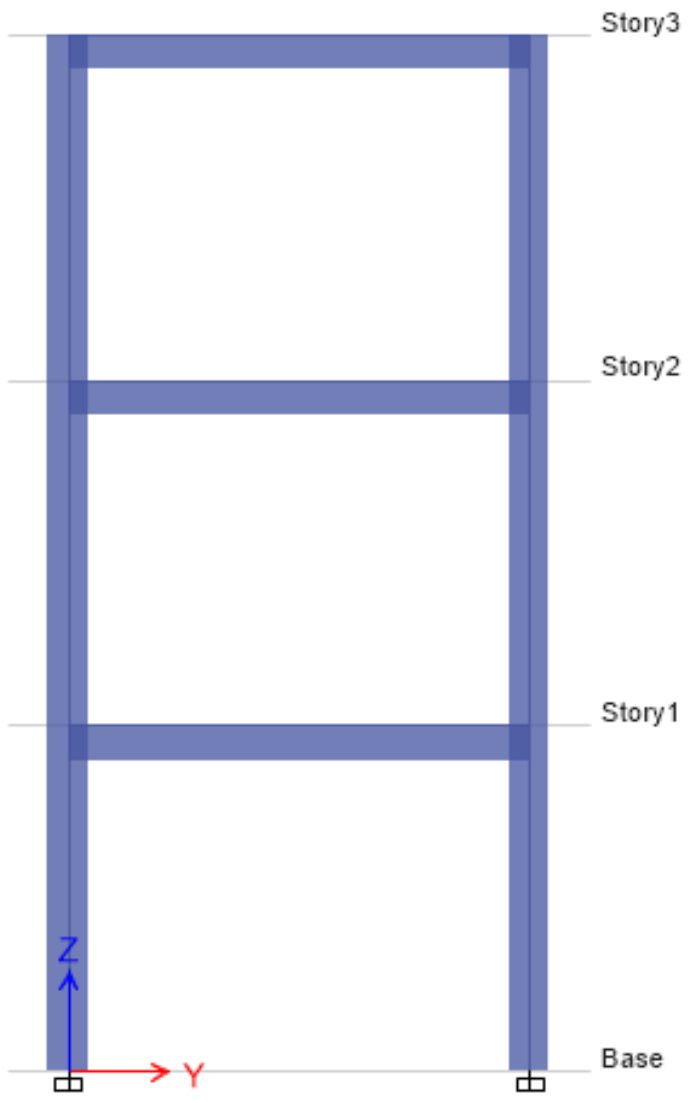

*Figure 5. Three-story RC frame (Monavari et al., 2018)*

Notably, this model was previously used for deterioration identification using a time-series and wavelet approach (Gharehbaghi et al., 2021; Gharehbaghi et al., 2020; Monavari et al., 2018). The input excitations were obtained from a real-world site under ambient environmental vibrations with a sampling rate of 200 Hz. The real site corresponds to the location of the P-block building at the Gardens Point campus of QUT, Australia. Since the input excitations are recorded from a real building site, they consist of a high noise level, human activities, traffic vibrations, and other environmental forces (Monavari, 2019).

In this case study, deterioration is defined as a continuous loss of cross-section. The structure was monitored for a 50-year period and comprising five states of intensity. During these states, the cross-section area of the left column's bars gradually reduced, and the annual deterioration rate (ADR) for the cross-section area was considered equal to $2\times10^{-3}$. The first state indicates the baseline (State 1), and the four following states are the deteriorated conditions starting from slight deterioration (State 2) to significant deterioration (State 5).

The deterioration modeling scenarios for the three cases are introduced in Table 1. Consequently, the deterioration rate (DR) for a year can be computed by multiplying the duration of deterioration (DOD) process and the ADR (Monavari et al., 2018).

$$ADR = \left\{1 - \frac{\text{Reduced cross - section area}}{\text{Refrenced cross - section area}}\right\} year \quad (2)$$

*Table 1. Deterioration scenarios (Monavari et al., 2018)*

| Scenario | Story 1 | Story 2 | Story 3 |
|---|---|---|---|
| 1 | Deteriorated | None | None |
| 2 | None | Deteriorated | None |
| 3 | None | None | Deteriorated |

***Second Case-Study: Damage Model***

Regarding the damage model, an SHM benchmark is presented here. This model is a three-story bookshelf with metal columns and aluminum floor plates, as shown in Figure 6. Moreover, four isolators allow the structure to sway in horizontal directions with the aid of a hydraulic shaker. The structure is instrumented by piezoelectric single-axis accelerometers. There are nine structural state conditions. The first state indicates the healthy state, and the rest are related to damage conditions. Damage and environmental changes were simulating by stiffness reduction of columns and replacing a mass, respectively. The healthy condition was considered with no stiffness reduction in sections and no mass on floors. For applying a damage scenario, columns were replaced by another section with a smaller section-area. A total of 450 records (nine states, each of which has 50 records) with a sampling rate of 320 Hz are recorded in each state, as shown in Table 2.

*Table 2. Data labels of structural states (T. Nguyen et al., 014)*

| Label | Record | Details | Description |
|---|---|---|---|
| State 1 | 0-50 | No mass and no stiffness reduction | Healthy |
| State 2 | 51-100 | Mass = 1.2 kg at the base | Environmental effects |
| State 3 | 101-150 | Mass = 1.2 kg on the 1st floor | Environmental effects |
| State 4 | 151-200 | 87.5% stiffness reduction in column 1BD | Damaged |
| State 5 | 201-250 | 87.5% stiffness reduction in column 1AD and 1BD | Damaged |
| State 6 | 251-300 | 87.5% stiffness reduction in column 2BD | Damaged |
| State 7 | 301-350 | 87.5% stiffness reduction in column 2AD and 2BD | Damaged |
| State 8 | 351-400 | 87.5% stiffness reduction in column 3BD | Damaged |
| State 9 | 401-450 | 87.5% stiffness reduction in column 3AD and 3BD | Damaged |

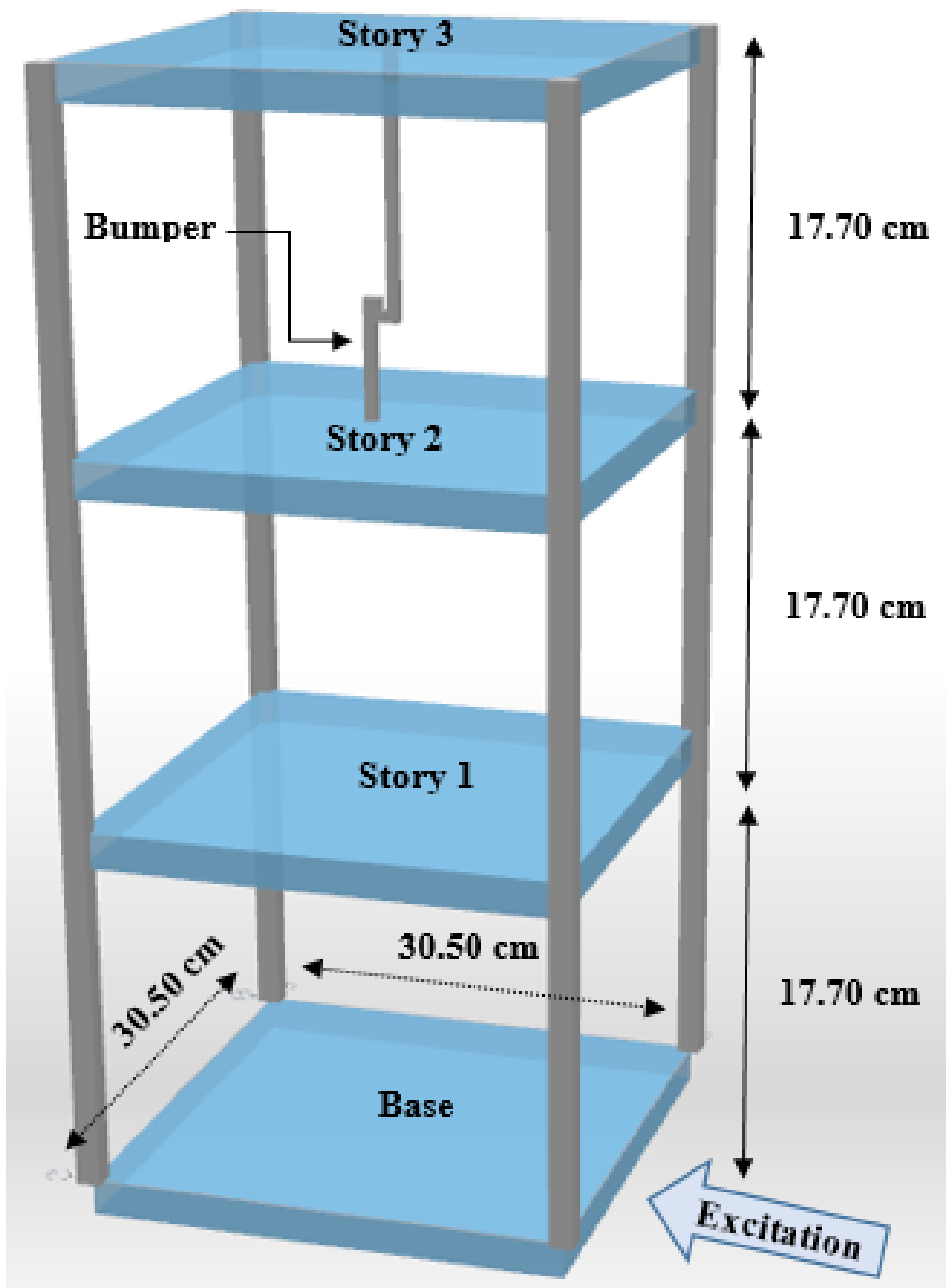

*Figure 6. Three-story bookshelf [adapted from (Gharehbaghi et al., 2021)]*

## Deterioration Identification Procedure (DIP)

Figure 7 presents the DIP procedure proposed. As shown, the proposed strategy is performed on the building dataset in three sequential steps, including pre-processing, processing in the time-frequency domain, and classification using CNN, which are explained in detail in the following sections.

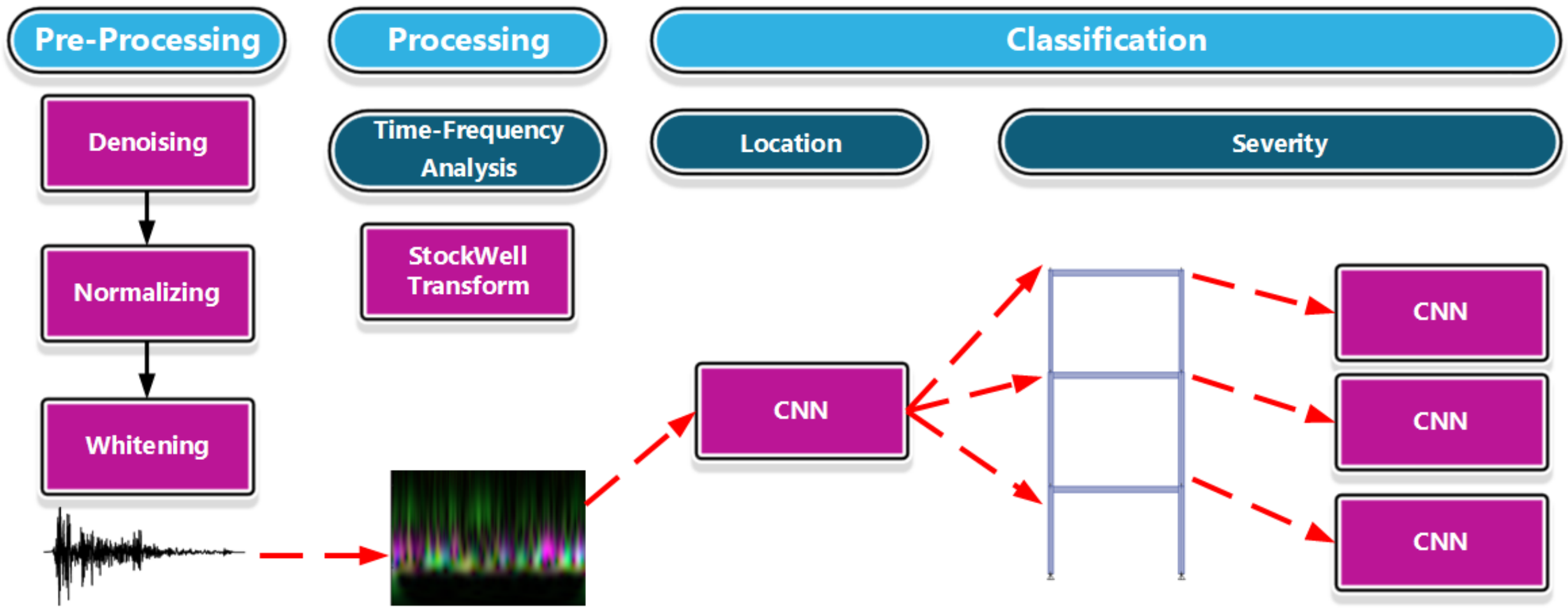

*Figure 7. DIP diagram*

***Step1: Pre-processing***

First, the high-frequency noise of the responses (story accelerations) is removed by employing a 12-order low-pass Chebyshev filter. This filter is selected because of the fast speed of denoising. More details about this filter can be found in (Vaseghi, 2008).

Next, mathematical and statistical operations are conducted on the data in order to reduce the effects of environmental and operational effects. Let $x_i$ and $\hat{x}_i$ be the measured and standardized signals; hence, the data normalization is carried out as follows (Monavari et al., 2018):

$$\hat{x}_i = \frac{x_i - \bar{x}}{\sigma} \tag{3}$$

where $\bar{x}$ and $\sigma$ show the mean and standard deviation of the measured signal.

Typically, in a building dataset, the responses of stories are correlated; hence, it is essential to conduct some operations for decorrelating or whitening. In this study, Zero-Phase Component Analysis (ZCA) is utilized to remove the cross-correlation between signals (Gharehbaghi et al., 2020). Afterward, the pre-processed signals are used in the next step in the time-frequency analysis.

***Step 2: Processing in Time-Frequency domain***

Several methods have been proposed and used for time-frequency analysis. Stockwell Transform (ST), introduced by Stockwell in 1996 (Kalbkhani & Shayesteh, 2017), shares similarities with STFT, and both techniques can reveal the time distribution of various frequency bands. Since STFT inherits a fixed window width, it cannot be implemented primarily for non-stationary signals.

By contrast, ST benefits from a flexible window as a frequency function, inheriting the benefits of the multi-resolution decomposition like continuous wavelet transform (CWT). Thus, while having higher frequency bands, it provides efficient processing capabilities with a fine-scaled window generating fine resolution and coarse resolution low frequencies (Kalbkhani & Shayesteh, 2017). Notably, ST can be presented using CWT and a phase factor. In this regard, the CWT of $x(t)$ is formulated as follows:

$$W(\tau,d)=\int_{-\infty}^{+\infty} x(t)\,w(t-\tau,d)dt \tag{4}$$

where $\tau$ and $d$ denote the location of the window at time axis and dilation term, respectively, and $w(t,d)$ is the mother wavelet defined as:

$$w(t,f)=\frac{|f|}{\sqrt{2\pi}}e^{-\frac{f^2t^2}{2}}e^{-j2\pi ft} \tag{5}$$

In the above, $f$ shows the frequency in Hz, which is inverse of $d$ $(f=1/d)$ and $j=\sqrt{-1}$. Accordingly, by multiplying the phase factor, the one-dimensional continuous ST of $x(t)$ is:

$$S(\tau,f)=W(\tau,d)e^{j2\pi f\tau} \tag{6}$$

Alternatively, by substituting (4), ST can also be written as follows (Sartipi et al., 2020):

$$S(\tau,f)=\frac{|f|}{\sqrt{2\pi}}\int_{-\infty}^{+\infty} x(t)\,e^{-\frac{(\tau-t)^2f^2}{2}}e^{-j2\pi ft}dt \tag{7}$$

Regarding (6) and (7), it is apparent that the window length varies. It means that, at low frequencies, the window becomes more expansive, while at high frequencies, it becomes narrower. Thus, ST is capable of providing a proper resolution within the various frequency bands of structural responses (Ghasemzadeh et al., 2018).

Moreover, ST has some merits compared to other transformations since (Stockwell, 2007):

- It can combine frequency-dependent resolution with an absolute reference phase.
- It measures the local amplitude spectrum coupled with the local phase spectrum, while WT provides local amplitude/power spectrum.
- Contrary to most of the WT approaches, ST can be implemented on complex time series.

Significantly, ST has been used mainly for medical diagnosis, including Magnetic resonance imaging (MRI) and electroencephalography (EEG) signals, and relatively few studies have been done in the realm of structural damage identification (Kalbkhani & Shayesteh, 2017; Sartipi et al., 2020). Accordingly, this study attempts to employ ST to identify building structures for the first time.

***Step3: Classification***

Similar to CWT and according to Eq. (7), the output of ST is complex-valued. Therefore, after analyzing each record of signals, a series of complex matrices are obtained. Considering the ST matrices, each from the record of one sensor, we have multi-dimensional features that can be represented in the form of spectrograms that are utilized as the input of robust CNNs. According to Figure 5, two types of CNNs are designed in this paper; one for localization and the other for assessing damage severity. First, a CNN is utilized on the colored images to detect the correct story level. Second, three distinctive CNNs classify different deterioration or damage scenarios.

Five-fold cross-validation is adopted for training and testing of CNNs in all classification problems. To this end, the available data is partitioned into five non-overlapping parts, and the training and testing procedure is repeated five times, and the results are then averaged. In each trial, four parts are used for training, and the remaining one is used for testing. The set of parameters used in the training process of CNNs are given in Table 3.

*Table 3. Parameters used in the training process*

| Parameter | Value |
|---|---|
| Optimizer | Stochastic gradient descent with momentum (SGDM) |
| Loss function | Cross-entropy |
| Momentum | 0.9 |
| Batch size | 32 |
| Maximum number of epochs | 200 |
| Learning Rate | 0.002 |
| Regularization parameter | 0.001 |
| Learning rate drop factor | 0.1 |
| Learning rate drop period | 20 |

Regarding the first case study and considering one sensor in each story, the input of localization CNN has a size of 256x1024x3. This model consists of 22 layers. It has three layers from each of convolution, batch normalization, max pooling, and fully connected layers. Also, there are five ReLU layers, two dropout layers, one softmax, and some classification layers. It should be mentioned that dropout layers are only used in the training process and are not involved in the test phase.

After locating the deteriorated story, the severity of deterioration is recognized using a different CNN. The CNN structure has three layers from each of the convolution, batch normalization, and max-pooling layers. Also, there are four ReLU layers, two fully connected layers, and one layer from each of the dropouts, softmax, and classification layers.

Regarding the second case study, a CNN with 20 layers is designed for localization of damage. After locating the damaged story, the severity is assessed using three distinctive CNNs designed for each story.

## Results and Discussion

In this section, the results and discussion on all case studies are given.

### *First Case Study (Deterioration)*

In the first case study, 1024 samples from deterioration signals are considered; as a result, the ST of the sensor record is a 513×1024 complex-valued matrix, and its absolute is utilized for the classification. There are three 513×1024 matrices for three sensors that can be depicted as colored images. The output of ST shows that there is no information in frequencies higher than 25 Hz in the deterioration signals since they are removed for the sake of better representation as well as reducing computational complexity. The ST of the healthy state and some of the deteriorated scenarios are depicted in Figure 8.

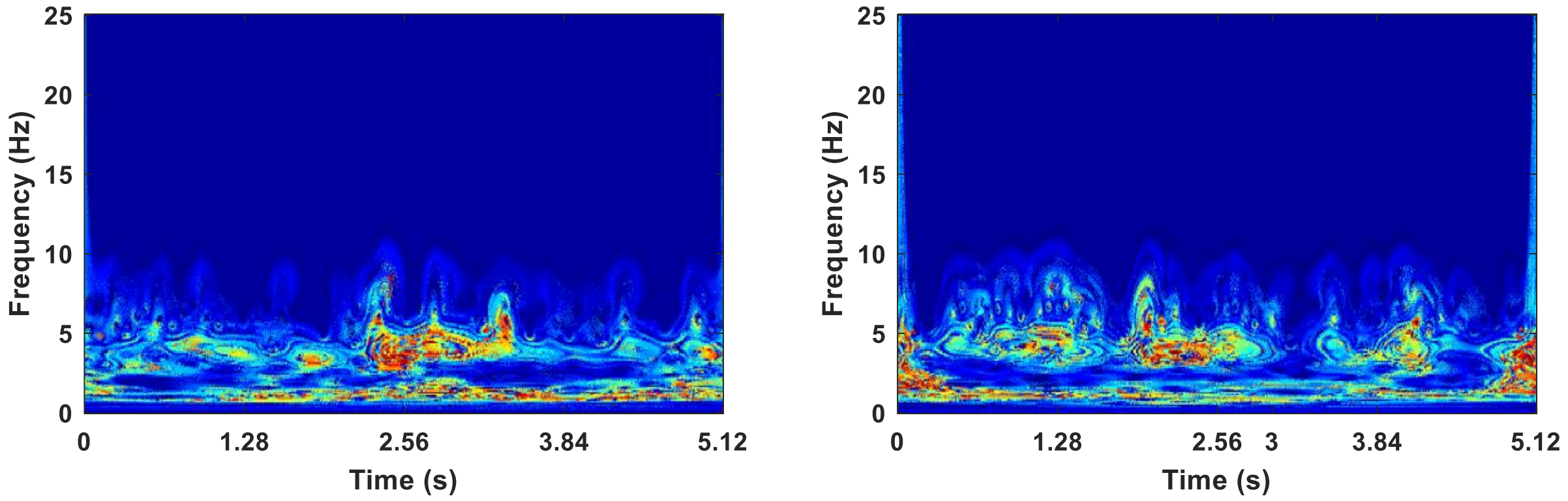

*(a) Healthy* *(b) Scenario 1, Level 1*

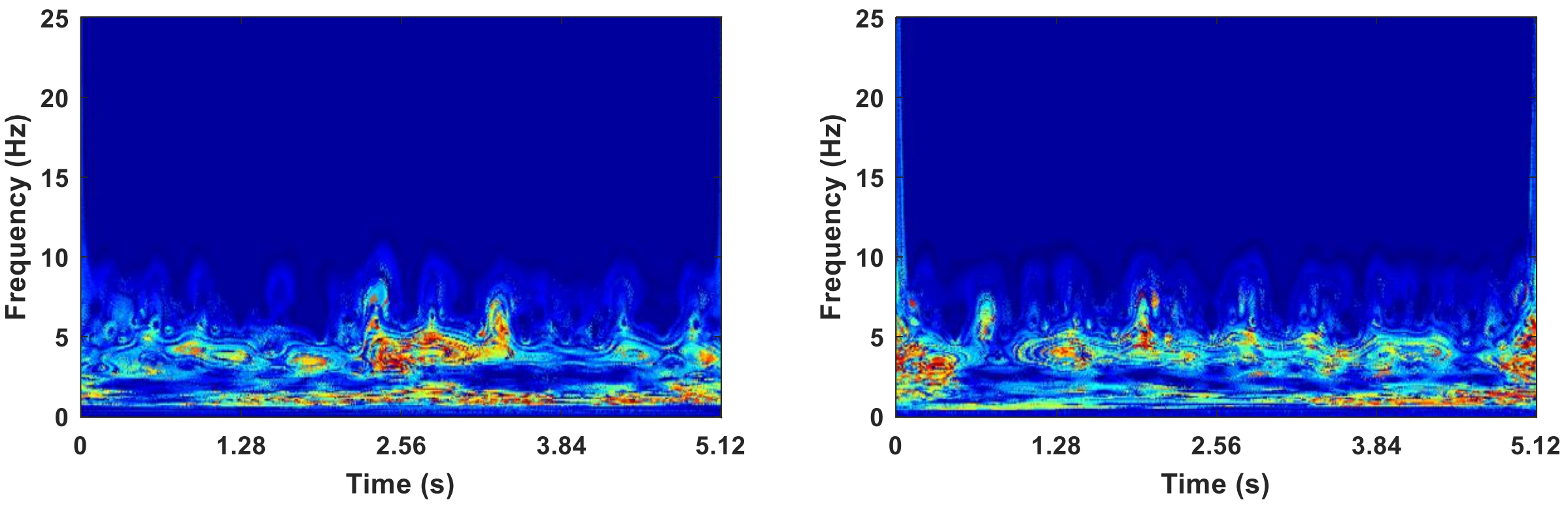

*(c) Scenario 2, Level 3* *(d) Scenario 3, Level 4*

*Figure 8. ST of the deterioration records.*

As mentioned before, deterioration localization and severity assessment are performed in two steps. At first, one CNN is designed for localization, and then one CNN is designed for identifying severity in each scenario. According to Table 1, there are three scenarios for localization of deterioration. These scenarios, along with the healthy (normal) class, form the 4-class classification problem to design the structure of localization CNN in the first case study. To this end, 80 records with a length of 1024 samples are considered from each class; therefore, a total of 320 records are used; 256 records for the training process, including 208 training data and 48 validation data, and 64 records as test data. The output of ST is a 513x1024 matrix covering the frequency range of [0 50] Hz. As mentioned, there is not discriminative information in frequencies higher than 25 Hz. After removing zero frequency and higher frequencies, the output of ST is 256x1024. For evaluating the method's performance, we computed three performance metrics as precision (Prec.), specificity (Spec.), and F1-score ($F_1$). These metrics are computed as:

$$\text{Prec.} = \frac{TP}{TP + FP} \tag{8}$$

$$\text{Spec.} = \frac{TN}{TN + FP} \tag{9}$$

$$F_1 = 2\,\frac{\text{Prec.} \times \text{Sens.}}{\text{Prec.} + \text{Sens.}} \tag{10}$$

where TP is the number of true-positive for the considered class. For a certain class, the true-positive is the signals that belong to this class, and the classifier correctly classifies. The false positive is the signals of other classes that are wrongly classified to the considered class.

TN is the true-negative, which is the number of signals that do not belong to the considered class, and the classifier does not classify them to the considered class. Also, Sens. is the sensitivity (or recall) which is computed as:

$$\text{Sen.} = \frac{TP}{TP + FN} \quad (11)$$

where FN is the number of false-negative predictions for the considered class, the false-negative is the signals that belong to the considered class, while the classifier is wrongly classified into other classes. Looking at the confusion matrix given in Table 4, the designed CNN can locate the affected story with an overall accuracy of 95%.

As discussed before, the deterioration severity is determined using a different CNN. Here, 20 records are supposed from each severity level at each level. Therefore, there are 80 records from each scenario; 68 records are used in the training process; 52 records as training data and 12 records as validation data, and 16 records are deployed as test data. Regarding confusion matrices depicted in Tables 5 to 7, it is evident that the minimum accuracy for all scenarios is more than 97%. Consequently, the proposed method can localize and qualify the deterioration precisely.

*Table 4. Confusion matrix for deterioration localization*

| Overall accuracy = 0.9470 | | Predicted class | | | | Sens. | Pre. | Spec. | $F_1$ |
|---|---|---|---|---|---|---|---|---|---|
| | | Healthy | Scenario 1 | Scenario 2 | Scenario 3 | | | | |
| Actual class | Healthy | 76 | 2 | 1 | 1 | 0.95 | 0.962 | 0.9875 | 0.9560 |
| | Scenario 1 | 1 | 77 | 2 | 1 | 0.9625 | 0.9277 | 0.975 | 0.9448 |
| | Scenario 2 | 1 | 2 | 75 | 2 | 0.9375 | 0.9493 | 0.9834 | 0.9434 |
| | Scenario 3 | 1 | 2 | 1 | 76 | 0.95 | 0.95 | 0.9834 | 0.9500 |
| Overall | | | | | | 0.9470 | 0.9472 | 0.9823 | 0.9486 |

*Table 5. Confusion matrix for deterioration severity (Scenario 1)*

| Overall accuracy 0.975 | | Predicted class | | | | Sens. | Pre. | Spec. | $F_1$ |
|---|---|---|---|---|---|---|---|---|---|
| | | State 2 | State 3 | State 4 | State 5 | | | | |
| Actual class | State 2 | 19 | 1 | 0 | 0 | 0.95 | 1.00 | 1.00 | 1.00 |
| | State 3 | 0 | 20 | 0 | 0 | 1.00 | 0.9090 | 0.9667 | 0.9370 |
| | State 4 | 0 | 1 | 19 | 0 | 0.95 | 1.00 | 1.00 | 1.00 |
| | State 5 | 0 | 0 | 0 | 20 | 1.00 | 1.00 | 1.00 | 1.00 |
| Overall | | | | | | 0.975 | 0.9772 | 0.9917 | 0.9843 |

*Table 6. Confusion matrix for deterioration severity (Scenario 2)*

| Overall accuracy = 0.9875 | | Predicted class | | | | Sens. | Pre. | Spec. | $F_1$ |
|---|---|---|---|---|---|---|---|---|---|
| | | State 2 | State 3 | State 4 | State 5 | | | | |
| Actual class | State 2 | 20 | 0 | 0 | 0 | 1.00 | 1.00 | 1.00 | 1.00 |
| | State 3 | 0 | 19 | 1 | 0 | 0.95 | 1.00 | 1.00 | 0.9744 |
| | State 4 | 0 | 0 | 20 | 0 | 1.00 | 0.9524 | 0.9833 | 0.9756 |
| | State 5 | 0 | 0 | 0 | 20 | 1.00 | 1.00 | 1.00 | 1.00 |
| Overall | | | | | | 0.9875 | 0.9881 | 0.9958 | 0.9875 |

*Table 7. Confusion matrix for deterioration severity (Scenario 3)*

| Overall accuracy = 0.975 | | Predicted class | | | | Sens. | Pre. | Spec. | $F_1$ |
|---|---|---|---|---|---|---|---|---|---|
| | | State 2 | State 3 | State 4 | State 5 | | | | |
| Actual class | State 2 | 20 | 0 | 0 | 0 | 1.00 | 1.00 | 1.00 | 1.00 |
| | State 3 | 0 | 20 | 0 | 0 | 1.00 | 0.9524 | 0.9833 | 0.9756 |
| | State 4 | 0 | 1 | 19 | 0 | 0.95 | 0.95 | 0.9833 | 0.9500 |
| | State 5 | 0 | 0 | 1 | 19 | 0.95 | 1.00 | 1.00 | 0.9744 |
| Overall | | | | | | 0.975 | 0.9756 | 0.9916 | 0.975 |

### *Second Case Study (Damage)*

Like the first case, 1024 samples from each record are considered to obtain the ST of damage signals. The results showed that there is not much useful information on frequencies above 80 Hz. Hence, they have been omitted for better representation and to reduce computational complexity. The ST of different states in the second case study is given in Figure 9.

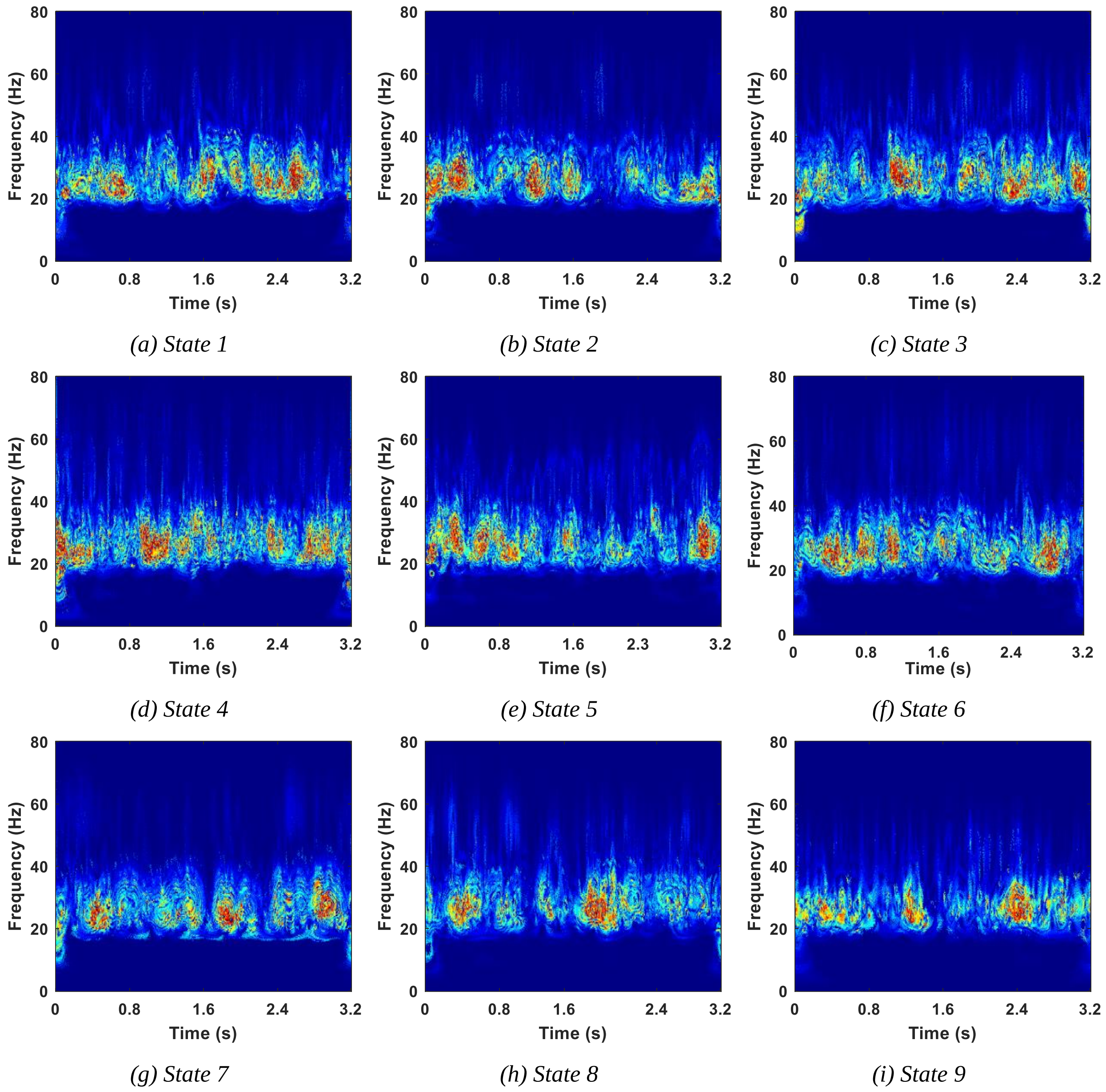

(a) State 1 (b) State 2 (c) State 3

(d) State 4 (e) State 5 (f) State 6

(g) State 7 (h) State 8 (i) State 9

*Fig. 9. ST of the damage records.*

A look through the ST of signals (Figures 8 and 9) discloses that the dominant frequency of damage sate is roughly 20 Hz, four times the deterioration with the frequency of 5 Hz. It should be noted that damage signals experience further fluctuations. Additionally, the damage signals have higher amplitudes since they are plotted with brighter colors coupled with higher contrast than the deterioration state.

Regarding the damage case, the same steps are taken, as illustrated for the deterioration case. Thus, according to Table 2, there are three scenarios for damage localization plus one healthy case. As a result, there are four classes in the classification problem.

As in the preceding case, 80 records with 1024 samples are considered for each class. A total of 320 records are used; 256 records for the training process, including 208 training data and 48 validation data, and 64 records for testing. The ST output is a 513x0124 matrix, including the frequency range of [0 80] Hz. Finally, after removing zero frequency and higher frequencies, the output of ST is 256x1024.

Conclusively, Table 8 denotes that the methodology's overall accuracy in locating damaged stories is roughly 96.0 percent. Eventually, as shown in confusion matrices (Tables 9-11), a minimum accuracy of about 97.0 percent is achieved by deploying the DIP on the damage case.

In the last section, the obtained results are compared with a similar work conducted by Gharehbaghi et al. (Gharehbaghi et al., 2021) named REF in this paper. In REF, the wavelet is used for signal processing in the time-frequency domain, and conventional machine learning algorithms are chosen for classification. To have a rational comparison, the accuracy in localization and severity determination are defined with an average index of both. For instance, regarding the first case study, in scenario 1, the overall accuracies for locating deteriorated story and severity assessments are 0.95 and 0.975, respectively. Therefore, the average accuracy is equal to 0.9625. The same procedure is conducted for calculating average accuracies in the REF. In Figure 10, the results for both methodologies are compared. It is evident that the DIP method has enhanced the performance of classification, especially for the second scenario, and the minimum accuracy of roughly 96% has been achieved in this paper.

*Table 8. Confusion matrix for damage localization*

| Overall accuracy = 0.982 | | Predicted class | | | | Sens. | Prec. | Spec. | $F_1$ |
|---|---|---|---|---|---|---|---|---|---|
| | | Healthy | Story 1 | Story 2 | Story 3 | | | | |
| Actual class | Healthy | 47 | 3 | 0 | 0 | 0.94 | 0.959 | 0.996 | 0.949 |
| | Story 1 | 2 | 198 | 0 | 0 | 0.99 | 0.971 | 0.976 | 0.98 |
| | Story 2 | 0 | 1 | 99 | 0 | 0.99 | 1 | 1 | 0.995 |
| | Story 3 | 0 | 2 | 0 | 98 | 0.98 | 1 | 1 | 0.989 |
| Overall | | | | | | 0.982 | 0.982 | 0.989 | 0.982 |

*Table 9. Confusion matrix for damage severity (Story 1)*

| Overall accuracy = 0.97 | | Predicted class | | | | Sens. | Prec. | Spec. | $F_1$ |
|---|---|---|---|---|---|---|---|---|---|
| | | State 2 | State 3 | State 4 | State 5 | | | | |
| Actual class | State 2 | 49 | 1 | 0 | 0 | 0.98 | 0.961 | 0.987 | 0.973 |
| | State 3 | 2 | 48 | 0 | 0 | 0.96 | 0.979 | 0.993 | 0.986 |
| | State 4 | 0 | 0 | 48 | 2 | 0.96 | 0.979 | 0.993 | 0.986 |
| | State 5 | 0 | 0 | 1 | 49 | 0.98 | 0.961 | 0.986 | 0.973 |
| Overall | | | | | | 0.97 | 0.97 | 0.989 | 0.979 |

*Table 10. Confusion matrix for damage severity (Story 2)*

| Overall accuracy = 0.99 | | Predicted class | | Sens. | Prec. | Spec. | $F_1$ |
|---|---|---|---|---|---|---|---|
| | | State 6 | State 7 | | | | |
| Actual class | State 6 | 49 | 1 | 0.98 | 1 | 1 | 1 |
| | State 7 | 0 | 50 | 1 | 0.98 | 0.98 | 0.98 |
| Overall | | | | 0.99 | 0.99 | 0.99 | 0.99 |

*Table 11. Confusion matrix for damage severity (Story 3)*

| Overall accuracy = 0.98 | | Predicted class | | Sens. | Prec. | Spec. | $F_1$ |
|---|---|---|---|---|---|---|---|
| | | Sate 8 | Sate 9 | | | | |
| Actual class | Sate 8 | 49 | 1 | 0.98 | 0.98 | 0.98 | 0.98 |
| | Sate 9 | 1 | 49 | 0.98 | 0.98 | 0.98 | 0.98 |
| Overall | | | | 0.98 | 0.98 | 0.98 | 0.98 |

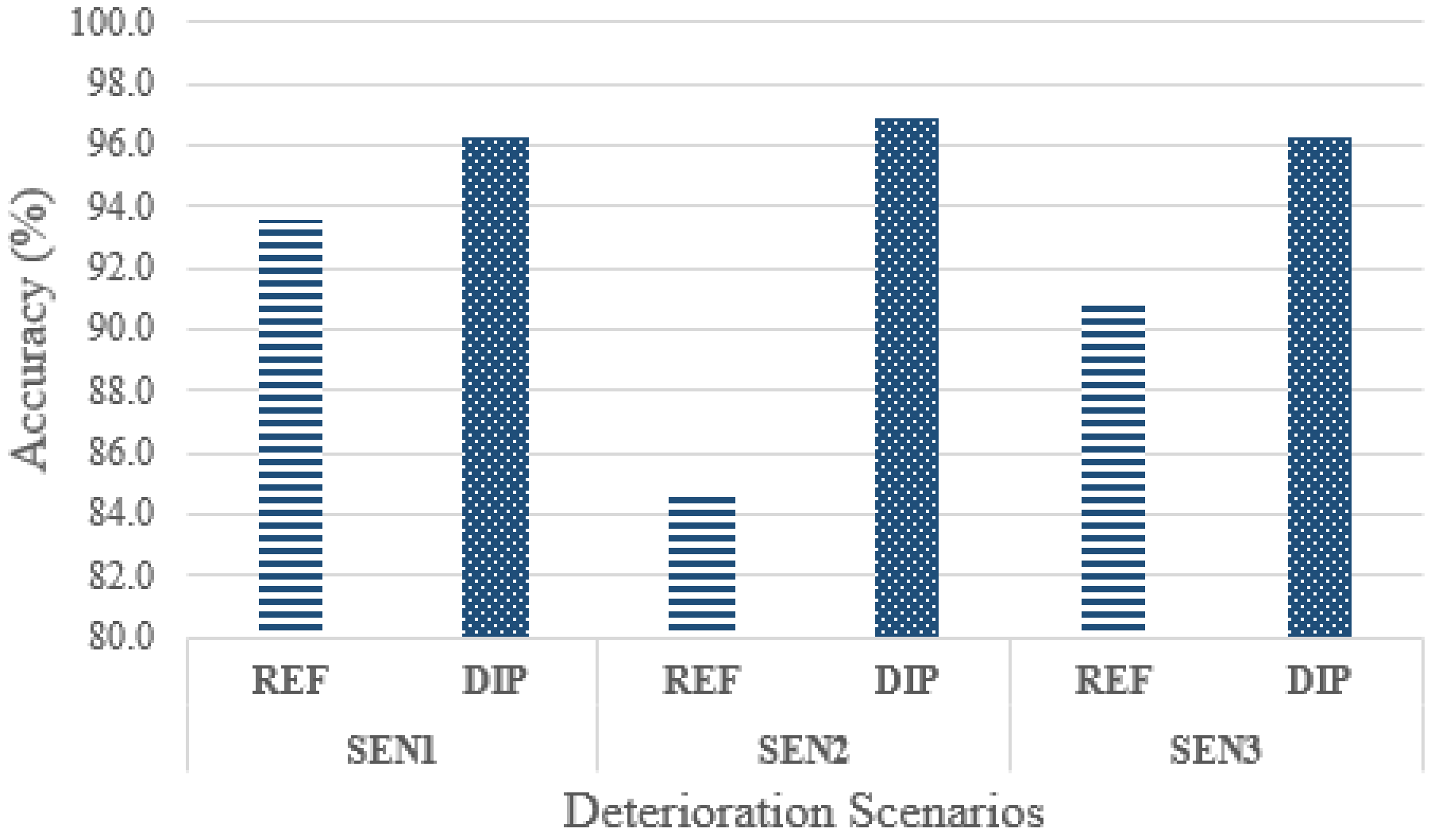

*Figure 10 Average accuracies for deterioration case*

Regarding the damage case study, the average results for REF and the DIP are denoted in Figure 11. Overall, the DIP reveals higher performance, particularly in the third story, from about 84% to near 97.0%. The figure proves that the minimum accuracy of the DIP is about 97.0% in detecting damage scenarios.

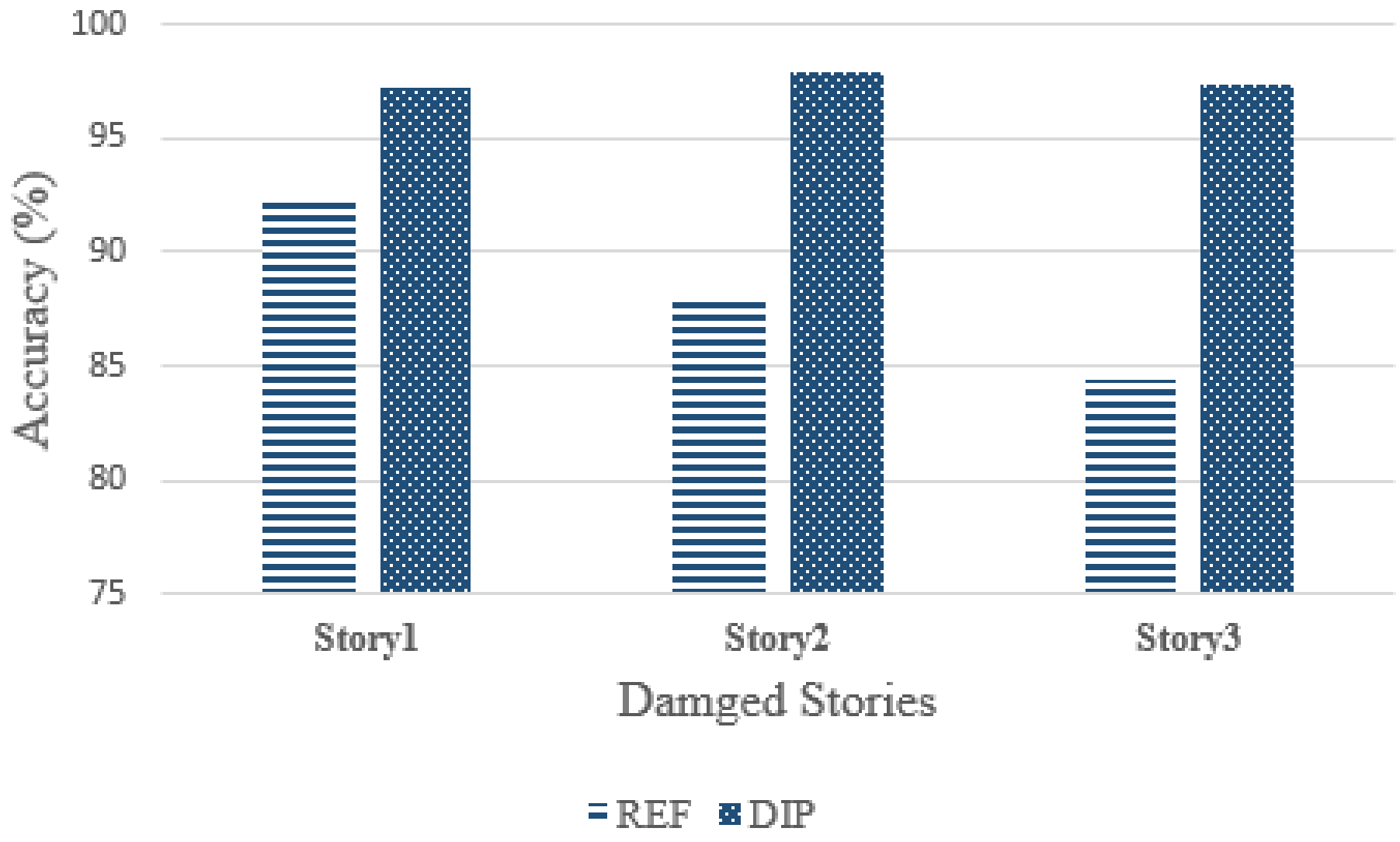

*Figure 11 Average accuracies for damage case*

## Conclusion

In this article, a novel methodology was presented using a data-driven approach capable of detecting deterioration and damage through output acceleration responses. To the best of the authors' knowledge, this is the first-ever development of a procedure that can detect both deterioration and damage via an innovative combination of Stockwell Transform (ST) and Convolutional Neural Networks (CNN). In this respect, an ST is applied to obtain complex matrices expressed as high dimensionality images representing various frequency bands in the time-frequency domain. After pre-processing the acceleration input, a CNN is employed to locate the affected story regarding the DIP. Afterward, another CNN is used to qualify damage and deterioration in each story. Notably, the established DIP successfully detected deterioration and damage in both models with high accuracy. Due to the flexibility of DIP, this procedure can also be utilized for various types of structures such as bridges or dams, along with various types of damage scenarios broadening the impact of this study. However, although the DIP has efficient performance regarding damage and deterioration detection under ambient and forced vibration with high accuracy, using separate CNNs for each story translates into high computational time to train the network. Thus, the authors are looking forward to establishing and enhancing DIP procedures that use only a CNN regardless of the number of stories.

## Disclosure statement

The author(s) declared no potential conflicts of interest concerning the research, authorship, and/or publication of this article.

## Replication of results

Implementation of the proposed numerical method was undertaken using MATLAB and IDARC platforms. All developed codes and models supporting the findings of this study are available from the corresponding author upon reasonable request.